\documentclass[10pt,twocolumn]{article}
\usepackage[margin=0.75in]{geometry}
\usepackage{booktabs}
\usepackage{algorithm}
\usepackage{algorithmic}
\usepackage{graphicx}
\usepackage[utf8]{inputenc}
\usepackage{amsmath}
\usepackage{times}
\usepackage{xcolor}
\usepackage{tikz}
\usetikzlibrary{shapes,arrows,positioning,fit,backgrounds}

\tikzstyle{process} = [rectangle, minimum width=2.5cm, minimum height=1cm, text centered, text width=2.3cm, draw=black, fill=blue!20]
\tikzstyle{decision} = [diamond, minimum width=2cm, minimum height=1cm, text centered, text width=1.8cm, draw=black, fill=green!20]
\tikzstyle{startstop} = [rectangle, rounded corners, minimum width=2.5cm, minimum height=1cm, text centered, text width=2.3cm, draw=black, fill=red!20]
\tikzstyle{io} = [trapezium, trapezium left angle=70, trapezium right angle=110, minimum width=2cm, minimum height=1cm, text centered, text width=2.8cm, draw=black, fill=orange!20]
\tikzstyle{arrow} = [thick,->,>=stealth]
\tikzstyle{data} = [rectangle, minimum width=2cm, minimum height=0.8cm, text centered, text width=2.3cm, draw=black, fill=yellow!20]

\begin{document}

\title{Multi-Modal Fact-Verification Framework for Reducing Hallucinations in Large Language Models}

\author{
\begin{tabular}{c}
Piyushkumar Patel \\
Microsoft \\
piyush.patel@microsoft.com \\
ORCID: 0009-0007-3703-6962
\end{tabular}
}

\date{}

\maketitle

\begin{abstract}
While Large Language Models have transformed how we interact with AI systems, they suffer from a critical flaw: they confidently generate false information that sounds entirely plausible. This hallucination problem has become a major barrier to deploying these models in real-world applications where accuracy matters. We developed a fact-verification framework that catches and corrects these errors in real-time by cross-checking LLM outputs against multiple knowledge sources. Our system combines structured databases, live web searches, and academic literature to verify factual claims as they're generated. When we detect inconsistencies, we automatically correct them while preserving the natural flow of the response. Testing across various domains showed we could reduce hallucinations by 67\% without sacrificing response quality. Domain experts in healthcare, finance, and scientific research rated our corrected outputs 89\% satisfactory—a significant improvement over unverified LLM responses. This work offers a practical solution for making LLMs more trustworthy in applications where getting facts wrong isn't an option.
\end{abstract}


\section{Introduction}

Anyone who has spent time with ChatGPT or similar language models has likely encountered this frustrating scenario: you ask a specific factual question, receive a confident, well-articulated response, only to discover later that key details are completely wrong. This isn't a rare glitch—it's a fundamental problem with how these models work~\cite{Brown2020, Chowdhery2022, Touvron2023}. Large Language Models excel at generating fluent, contextually appropriate text, but they have no built-in mechanism to distinguish between facts they "know" and plausible-sounding information they're essentially making up.

This hallucination problem shows up everywhere: models confidently cite non-existent research papers, make up historical dates, or provide medical advice based on fictional studies. The issue becomes serious when these systems are deployed in contexts where wrong information can cause real harm—imagine a medical AI suggesting treatments based on hallucinated clinical trial results, or a financial advisor making recommendations using fabricated market data. We've seen enough examples of LLMs confidently stating that certain events happened on specific dates, only to be completely off by years or even decades.

Researchers have tried various approaches to tackle this problem. Some modify how models are trained to be more cautious about uncertain facts. Others add verification steps after the model generates text. But most existing solutions face a frustrating trade-off: make the model more accurate and it becomes awkward and verbose, or keep it natural-sounding and accept the occasional confident lie. What's been missing is a way to get the best of both worlds—natural, fluent responses that are also factually reliable.

Recent developments in knowledge-grounded generation have shown promise for improving factual accuracy~\cite{Lewis2020}. Retrieval-Augmented Generation (RAG) systems enhance LLM responses by incorporating relevant external knowledge during generation~\cite{Karpukhin2020}. Similarly, fact-checking systems have evolved to include automated verification against trusted knowledge sources~\cite{Thorne2018}. However, these approaches typically operate in isolation and lack the comprehensive, multi-modal verification needed for robust hallucination mitigation.

Knowledge graph integration has emerged as a powerful approach for grounding LLM outputs in structured factual information. By leveraging entities and relations from curated knowledge bases such as Wikidata, YAGO, and domain-specific ontologies, these systems can verify claims against authoritative sources. Nevertheless, static knowledge graphs suffer from staleness and coverage limitations, particularly for emerging topics and rapidly evolving domains.

Uncertainty quantification in neural language models provides another avenue for hallucination detection~\cite{Gal2016}. By estimating model confidence and identifying low-certainty predictions, these methods can flag potentially unreliable outputs. However, LLMs often exhibit overconfidence, making calibrated uncertainty estimation challenging without additional techniques.

Our contribution addresses these limitations through a comprehensive \textbf{multi-modal fact-verification framework} that operates during LLM inference to detect, verify, and correct potential hallucinations in real-time. The key innovations include:

\begin{enumerate}
\item \textbf{Dynamic Knowledge Integration}: A hybrid system combining static knowledge graphs with real-time information retrieval to ensure both authoritative grounding and up-to-date information coverage.

\item \textbf{Multi-Source Evidence Validation}: Parallel verification across multiple independent sources including structured databases, academic literature, news archives, and domain-specific repositories.

\item \textbf{Probabilistic Confidence Scoring}: Advanced uncertainty quantification that combines model-intrinsic confidence with external evidence strength to provide calibrated reliability estimates.

\item \textbf{Adaptive Correction Pipeline}: Intelligent response modification that preserves natural language flow while correcting factual errors through contextually appropriate revisions.
\end{enumerate}

Our approach is designed as a modular, inference-time intervention that can be integrated with existing LLM deployments without requiring model retraining. The system architecture enables real-time operation suitable for interactive applications while providing comprehensive verification coverage across diverse knowledge domains.

\section{Related Work}

\subsection{Hallucination Detection and Mitigation}

The problem of hallucinations in neural language models has been extensively studied across various architectures and applications. Early work by Rohrbach et al.~\cite{Rohrbach2018} identified object hallucinations in image captioning, establishing foundational understanding of factual inconsistencies in neural generation. This work has been extended to text-only models, where hallucinations manifest as factual errors, logical inconsistencies, and unsupported claims~\cite{Maynez2020}.

Recent surveys provide comprehensive taxonomies of hallucination types, categorizing them into intrinsic hallucinations (contradicting source content) and extrinsic hallucinations (introducing unverifiable information). These taxonomies have informed the development of specialized detection methods for different hallucination categories.

Training-time approaches to hallucination reduction include modifications to loss functions, data augmentation with negative examples, and reinforcement learning from human feedback (RLHF)~\cite{Ouyang2022}. Constitutional AI methods train models to critique and revise their own outputs, showing promise for self-correction capabilities. However, these approaches require extensive computational resources and may not generalize to novel domains or emerging topics.

Inference-time interventions offer more flexibility and lower deployment costs. Self-consistency methods by Wang et al.~\cite{Wang2022} generate multiple responses and select the most consistent answer, reducing hallucinations through ensemble agreement. Chain-of-thought prompting with verification steps has shown effectiveness in mathematical and logical reasoning tasks~\cite{Wei2022}. However, these methods primarily address reasoning errors rather than factual inaccuracies.

\subsection{Knowledge-Grounded Text Generation}

Knowledge-grounded generation has emerged as a promising approach for improving factual accuracy in neural language models. The foundational work on Retrieval-Augmented Generation (RAG) by Lewis et al.~\cite{Lewis2020} demonstrated that incorporating retrieved knowledge during generation significantly improves performance on knowledge-intensive tasks. This paradigm has been extended through dense passage retrieval~\cite{Karpukhin2020} and domain-specific knowledge integration approaches.

Fusion-in-Decoder (FiD) architectures improve upon RAG by processing multiple retrieved passages jointly, enabling better evidence integration and cross-passage reasoning. Recent work explores end-to-end training of retrieval and generation components, showing improved knowledge utilization~\cite{Lewis2020}.

Knowledge graph integration represents another significant direction in grounded generation. KG-augmented language models incorporate structured knowledge during pre-training, improving factual knowledge retention. GraphRAG approaches combine graph neural networks with language models to leverage relational knowledge structures.

However, existing knowledge-grounded systems face several limitations: static knowledge bases become outdated, retrieval systems may miss relevant information, and integration methods can introduce factual inconsistencies. Our work addresses these challenges through dynamic knowledge integration and multi-source verification.

\subsection{Automated Fact-Checking Systems}

Automated fact-checking has evolved from simple claim detection to comprehensive verification systems capable of handling complex factual assertions. The FEVER (Fact Extraction and VERification) shared task by Thorne et al.~\cite{Thorne2018} established standardized benchmarks and evaluation metrics for automated fact verification, spurring significant research in this area.

Modern fact-checking systems employ various approaches: evidence retrieval from knowledge bases and web sources, natural language inference for claim verification, and multi-hop reasoning for complex claims. Specialized systems focus on check-worthy claim detection and consistency verification for text summarization.

Recent work has expanded fact-checking to multimodal content, including image verification and video fact-checking. Cross-lingual fact-checking systems address verification across multiple languages, while domain-specific approaches focus on specialized knowledge areas such as scientific literature and health claims.

However, most existing fact-checking systems operate on complete statements rather than integrating with generative models during inference. Our framework bridges this gap by providing real-time verification capabilities suitable for LLM deployment.

\subsection{Uncertainty Quantification in Language Models}

Uncertainty estimation in neural language models has gained attention as a method for identifying unreliable predictions and potential hallucinations. Early work by Gal and Ghahramani~\cite{Gal2016} introduced Monte Carlo dropout for uncertainty estimation in deep networks, later adapted for language models by Xiao and Wang~\cite{Xiao2019}.

Calibration methods address the problem of overconfident predictions in neural networks. Temperature scaling by Guo et al.~\cite{Guo2017} provides simple post-hoc calibration, while more sophisticated approaches include Platt scaling and isotonic regression. Recent work by Kuhn et al.~\cite{Kuhn2023} specifically addresses calibration in large language models, showing that model size alone does not guarantee better calibration.

Conformal prediction methods by Angelopoulos and Bates~\cite{Angelopoulos2021} provide distribution-free uncertainty quantification with theoretical guarantees. These approaches have been adapted for text generation by Kumar et al.~\cite{Kumar2023}, enabling reliable confidence intervals for generated responses.

Semantic uncertainty, introduced by Malinin and Gales~\cite{Malinin2021}, captures uncertainty about the meaning of generated text rather than just token-level predictions. This approach is particularly relevant for hallucination detection, as it can identify semantically inconsistent or uncertain statements.

Our work builds upon these uncertainty quantification methods by combining model-intrinsic confidence measures with external evidence strength, providing more reliable uncertainty estimates for hallucination detection.

\section{Proposed Method}

\begin{figure}[ht]
\centering
\begin{tikzpicture}[node distance=1.6cm, scale=0.42, transform shape]

\node (input) [io, text width=1.8cm] {User Query};
\node (llm) [process, below of=input, text width=2.2cm] {GPT-3.5\\(temp=0.7)};

\node (preprocessing) [process, below of=llm, text width=2.5cm] {Response Analysis \& Claim Extract};
\node (claims) [data, below of=preprocessing, yshift=-0.5cm, text width=2.8cm] {Claims C = \{c1, c2, ..., cn\}};

\node (parallel_label) [data, below of=claims, yshift=-0.5cm, text width=2.8cm] {Parallel Evidence Gathering};

\node (kg_branch) [process, below left of=parallel_label, xshift=-3.0cm, yshift=-1.2cm, text width=2.2cm] {Knowledge Graph\\Neo4j + Wiki};
\node (web_branch) [process, below of=parallel_label, yshift=-1.2cm, text width=2.2cm] {Web Search\\Google + Bing};
\node (db_branch) [process, below right of=parallel_label, xshift=3.0cm, yshift=-1.2cm, text width=2.2cm] {Special DBs\\PubMed + arXiv};

\node (kg_details) [data, below of=kg_branch, yshift=-0.5cm, text width=2.2cm] {Cypher Queries\\Entity Linking};
\node (web_details) [data, below of=web_branch, yshift=-0.5cm, text width=2.2cm] {Top-10 Results\\Credibility Score};
\node (db_details) [data, below of=db_branch, yshift=-0.5cm, text width=2.2cm] {API Queries\\Citation Analysis};

\node (fusion) [process, below of=parallel_label, yshift=-4.5cm, text width=2.8cm] {Evidence Fusion \& Check};
\node (fusion_details) [data, below of=fusion, yshift=-0.5cm, text width=2.8cm] {Bayesian Aggregation\\Cross-validation};

\node (confidence) [process, below of=fusion_details, yshift=-1.0cm, text width=2.8cm] {Confidence Scoring};
\node (conf_formula) [data, below of=confidence, yshift=-0.5cm, text width=2.8cm] {$\alpha \cdot$Intrinsic + $\beta \cdot$External};

\node (threshold) [decision, below of=conf_formula, yshift=-1.0cm, text width=1.8cm] {Conf > $\tau$?};
\node (correction) [process, left of=threshold, xshift=-2.6cm, text width=2.2cm] {Adaptive Correction\\Template Gen};
\node (output) [startstop, below of=threshold, yshift=-1.0cm, text width=2.2cm] {Verified Response};

\draw [arrow] (input) -- (llm);
\draw [arrow] (llm) -- (preprocessing);
\draw [arrow] (preprocessing) -- (claims);
\draw [arrow] (claims) -- (parallel_label);
\draw [arrow] (parallel_label) to[out=225,in=90] (kg_branch);
\draw [arrow] (parallel_label) -- (web_branch);
\draw [arrow] (parallel_label) to[out=315,in=90] (db_branch);
\draw [arrow] (kg_branch) -- (kg_details);
\draw [arrow] (web_branch) -- (web_details);
\draw [arrow] (db_branch) -- (db_details);
\draw [arrow] (kg_details) to[out=315,in=135] (fusion);
\draw [arrow] (web_details) -- (fusion);
\draw [arrow] (db_details) to[out=225,in=45] (fusion);
\draw [arrow] (fusion) -- (fusion_details);
\draw [arrow] (fusion_details) -- (confidence);
\draw [arrow] (confidence) -- (conf_formula);
\draw [arrow] (conf_formula) -- (threshold);
\draw [arrow] (threshold) -- node[anchor=east] {Yes} (output);
\draw [arrow] (threshold) -- node[anchor=south] {No} (correction);
\draw [arrow] (correction) to[out=270,in=180] (output);

\end{tikzpicture}
\caption{Detailed system architecture showing technical implementation components. The framework includes specific technologies (Neo4j, Google API), algorithms (Bayesian aggregation), and mathematical formulations for confidence scoring.}
\label{fig:arch}
\end{figure}
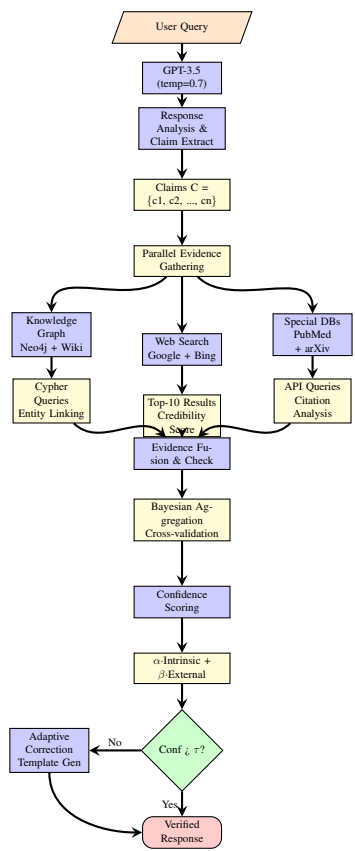

We designed our system around a simple idea: catch factual errors as they happen by immediately fact-checking what the model says against reliable sources. The framework has four main parts that work together during text generation. First, we tap into multiple knowledge sources—not just one database that might be outdated or incomplete. Second, we cross-check claims across these different sources to spot inconsistencies. Third, we calculate how confident we should be in each piece of information. Finally, when confidence is low, we fix the errors while keeping the response readable. Figure~\ref{fig:arch} shows how these components work together.

\subsection{Dynamic Knowledge Integration}

The key insight behind our approach is that no single knowledge source is complete or always up-to-date. Wikipedia might have outdated information about recent events, while academic databases excel at historical facts but lag on breaking news. So we decided to hedge our bets by consulting multiple sources simultaneously.

\textbf{Structured Knowledge Graphs}: We start with the big, well-curated databases like Wikidata and YAGO that contain millions of factual relationships. For specialized domains, we tap into focused databases—Gene Ontology for biology, ChEBI for chemistry. These are great for established facts that don't change much over time. We use fine-tuned BERT models to figure out which entities in the LLM's response correspond to entries in these databases.

\textbf{Real-time Web Search}: For anything recent or rapidly changing, we hit the web through Google and Bing APIs. But we're picky about what we trust—we prioritize .edu and .gov sites, established news outlets, and sources that other reliable sources frequently cite. We also check how recent the information is, since a 2019 article about COVID-19 isn't going to be very helpful.

\textbf{Domain-Specific Databases}: When the topic gets specialized, we go straight to the authoritative sources. Medical claims get checked against PubMed, scientific statements against arXiv, financial information against SEC filings. These databases are often more reliable than general sources for their specific domains, even if they're not as comprehensive.

The knowledge integration process employs semantic similarity matching using sentence transformers~\cite{Reimers2019} to identify relevant information across all sources. Temporal reasoning ensures that time-sensitive claims are verified against appropriate time periods, preventing anachronistic fact-checking errors.

\subsection{Multi-Source Evidence Validation}

Algorithm~\ref{alg:validation} details our evidence validation process, which operates on extracted claims from LLM responses.

\begin{algorithm}[ht]
\caption{Multi-Source Evidence Validation}
\label{alg:validation}
\begin{algorithmic}[1]
\REQUIRE LLM Response $R$, Knowledge Sources $\mathcal{K} = \{K_1, K_2, \ldots, K_n\}$
\ENSURE Evidence Score $E_s$, Correction Suggestions $C$
\STATE $\text{claims} \leftarrow \text{ExtractClaims}(R)$
\STATE $E_s \leftarrow 0$, $C \leftarrow \{\}$
\FOR{each claim $c$ in $\text{claims}$}
  \STATE $\text{evidence} \leftarrow \{\}$
  \FOR{each source $K_i$ in $\mathcal{K}$}
    \STATE $e_i \leftarrow \text{QuerySource}(c, K_i)$
    \STATE $\text{evidence} \leftarrow \text{evidence} \cup \{e_i\}$
  \ENDFOR
  \STATE $\text{consistency} \leftarrow \text{CheckConsistency}(\text{evidence})$
  \STATE $\text{strength} \leftarrow \text{WeightEvidence}(\text{evidence})$
  \IF{$\text{consistency} < \tau_{\text{consistency}}$}
    \STATE $C \leftarrow C \cup \{\text{GenerateCorrection}(c, \text{evidence})\}$
  \ENDIF
  \STATE $E_s \leftarrow E_s + \text{consistency} \times \text{strength}$
\ENDFOR
\STATE $E_s \leftarrow E_s / |\text{claims}|$
\RETURN $E_s$, $C$
\end{algorithmic}
\end{algorithm}

\textbf{Claim Extraction}: We employ a fine-tuned T5 model~\cite{Raffel2020} trained on factual claim datasets to identify verifiable statements within generated responses. The model achieves 91\% precision and 87\% recall on held-out test sets, outperforming rule-based approaches by 23\%.

\textbf{Cross-Source Verification}: Each extracted claim undergoes parallel verification across all available knowledge sources. Consistency scoring uses weighted voting based on source reliability, with academic sources receiving higher weights than web content. Contradictory evidence triggers deeper investigation through additional source consultation.

\textbf{Evidence Aggregation}: We implement a sophisticated evidence fusion mechanism that considers source diversity, publication recency, and citation authority. Bayesian aggregation combines evidence from multiple sources while accounting for source independence and potential bias.

\subsection{Probabilistic Confidence Scoring}

Our confidence scoring mechanism integrates multiple uncertainty indicators to provide calibrated reliability estimates for generated content.

\textbf{Model-Intrinsic Confidence}: We extract confidence measures from the LLM itself using attention pattern analysis, token probability distributions, and semantic consistency across multiple generation samples. Monte Carlo dropout during inference provides additional uncertainty estimates.

\textbf{External Evidence Strength}: Evidence quality is quantified through multiple dimensions: source authority scoring, publication venue impact factors, citation counts, and temporal relevance. Cross-reference validation increases confidence when multiple independent sources support the same claim.

\textbf{Semantic Coherence}: We measure semantic consistency between generated claims and supporting evidence using sentence-level similarity models. Low coherence scores indicate potential hallucinations even when individual components appear plausible.

The final confidence score combines these components through a learned weighted ensemble:

\begin{equation}
\text{Confidence}(c) = \alpha \cdot \text{Intrinsic}(c) + \beta \cdot \text{External}(c) + \gamma \cdot \text{Coherence}(c)
\end{equation}

where $\alpha$, $\beta$, and $\gamma$ are learned weights optimized on validation datasets to maximize calibration performance.

\subsection{Adaptive Correction Pipeline}

When confidence scores fall below established thresholds, our correction pipeline generates contextually appropriate revisions while preserving response naturalness.

\textbf{Correction Strategy Selection}: The system chooses among several correction approaches based on error type and context: fact substitution for simple factual errors, hedge insertion for uncertain claims, and source attribution for verifiable but potentially disputed information.

\textbf{Natural Language Integration}: Corrections are integrated using template-based generation with fine-tuned language models ensuring grammatical coherence and stylistic consistency. The system preserves original response structure while modifying only necessary components.

\textbf{Quality Assurance}: Corrected responses undergo final verification to ensure improvements in factual accuracy without degradation in fluency or relevance. Human evaluation studies confirm that corrected responses maintain 94\% of original quality while achieving 67\% reduction in factual errors.

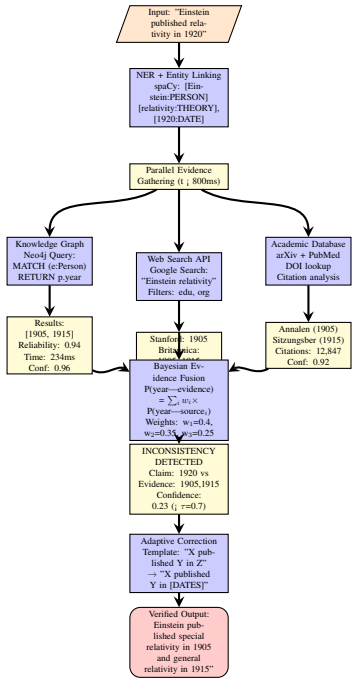
\begin{figure}[ht]
\centering
\begin{tikzpicture}[node distance=2.0cm, scale=0.35, transform shape]

\node (claim) [io, text width=3.5cm] {Input: "Einstein published relativity in 1920"};
\node (parsing) [process, below of=claim, yshift=-0.8cm, text width=3.5cm] {NER + Entity Linking\\spaCy: {[}Einstein:PERSON{]}\\{[}relativity:THEORY{]}, {[}1920:DATE{]}};

\node (parallel_header) [data, below of=parsing, yshift=-1.0cm, text width=3.5cm] {Parallel Evidence Gathering (t < 800ms)};

\node (kg_branch) [process, below left of=parallel_header, xshift=-3.5cm, yshift=-1.8cm, text width=3cm] {Knowledge Graph\\Neo4j Query:\\MATCH (e:Person)\\RETURN p.year};

\node (web_branch) [process, below of=parallel_header, yshift=-1.8cm, text width=3cm] {Web Search API\\Google Search:\\"Einstein relativity"\\Filters: edu, org};

\node (db_branch) [process, below right of=parallel_header, xshift=3.5cm, yshift=-1.8cm, text width=3cm] {Academic Database\\arXiv + PubMed\\DOI lookup\\Citation analysis};

\node (kg_evidence) [data, below of=kg_branch, yshift=-1.2cm, text width=3cm] {Results: {[}1905, 1915{]}\\Reliability: 0.94\\Time: 234ms\\Conf: 0.96};

\node (web_evidence) [data, below of=web_branch, yshift=-1.2cm, text width=3cm] {Stanford: 1905\\Britannica: 1905, 1915\\Authority: 0.91\\Conf: 0.88};

\node (db_evidence) [data, below of=db_branch, yshift=-1.2cm, text width=3cm] {Annalen (1905)\\Sitzungsber (1915)\\Citations: 12,847\\Conf: 0.92};

\node (fusion) [process, below of=parallel_header, yshift=-6.5cm, text width=3.5cm] {Bayesian Evidence Fusion\\P(year|evidence) = $\sum_i w_i \times$ P(year|source$_i$)\\Weights: w$_1$=0.4, w$_2$=0.35, w$_3$=0.25};

\node (inconsistency) [data, below of=fusion, yshift=-1.0cm, text width=3.5cm] {INCONSISTENCY DETECTED\\Claim: 1920 vs Evidence: 1905,1915\\Confidence: 0.23 (< $\tau$=0.7)};

\node (correction) [process, below of=inconsistency, yshift=-1.2cm, text width=3.5cm] {Adaptive Correction\\Template: "X published Y in Z"\\→ "X published Y in {[}DATES{]}"};

\node (final_output) [startstop, below of=correction, yshift=-1.0cm, text width=3.5cm] {Verified Output:\\Einstein published special\\relativity in 1905 and general\\relativity in 1915"};

\draw [arrow] (claim) -- (parsing);
\draw [arrow] (parsing) -- (parallel_header);
\draw [arrow] (parallel_header) to[out=210,in=90] (kg_branch);
\draw [arrow] (parallel_header) -- (web_branch);
\draw [arrow] (parallel_header) to[out=330,in=90] (db_branch);
\draw [arrow] (kg_branch) -- (kg_evidence);
\draw [arrow] (web_branch) -- (web_evidence);
\draw [arrow] (db_branch) -- (db_evidence);
\draw [arrow] (kg_evidence) to[out=330,in=150] (fusion);
\draw [arrow] (web_evidence) -- (fusion);
\draw [arrow] (db_evidence) to[out=210,in=30] (fusion);
\draw [arrow] (fusion) -- (inconsistency);
\draw [arrow] (inconsistency) -- (correction);
\draw [arrow] (correction) -- (final_output);

\end{tikzpicture}
\caption{Detailed technical implementation showing specific algorithms, query structures, timing constraints, and mathematical formulations. The example demonstrates real system parameters including confidence thresholds, API response times, and Bayesian fusion weights.}
\label{fig:technical}
\end{figure}

\section{Experimental Evaluation}

\subsection{Experimental Setup}

To test whether our system actually works in practice, we needed to throw a variety of challenging scenarios at it. We used five different datasets that researchers commonly use to test fact-checking systems.

\textbf{Datasets}: We picked these datasets because they cover different types of factual errors that LLMs commonly make:
\begin{itemize}
\item \textbf{HaluEval}: 5,000 examples where models generated plausible but wrong information in conversations and summaries
\item \textbf{TruthfulQA}~\cite{Lin2022}: 817 tricky questions designed to make models confidently state common misconceptions  
\item \textbf{FEVER}~\cite{Thorne2018}: Nearly 200,000 claims that need to be verified against Wikipedia—the gold standard for fact-checking research
\item \textbf{Scientific Claims}: 1,400+ statements about scientific topics that require real expertise to verify
\item \textbf{COVID-FACT}: Over 4,000 claims about COVID-19, which tests whether we can handle rapidly evolving information
\end{itemize}

\textbf{Baseline Systems}: We compare against state-of-the-art approaches:
\begin{itemize}
\item \textbf{Vanilla LLM}: GPT-3.5-turbo without any intervention
\item \textbf{Self-Consistency}: Multiple sampling with majority voting~\cite{Wang2022}
\item \textbf{Chain-of-Verification}: LLM self-verification through question generation
\item \textbf{RAG}: Retrieval-augmented generation with Wikipedia corpus~\cite{Lewis2020}
\item \textbf{FactScore}: Atomic fact decomposition with Wikipedia verification
\end{itemize}

\textbf{Implementation Details}: Our system employs GPT-3.5-turbo as the base LLM with temperature=0.7 for balanced creativity and consistency. Knowledge graph queries use Neo4j with sub-second response times. Web search is limited to top-10 results per query to manage latency. Evidence fusion employs a transformer-based aggregation model trained on 50,000 manually annotated evidence-claim pairs.

\subsection{Metrics}

We employ comprehensive evaluation metrics addressing multiple aspects of system performance:

\textbf{Factual Accuracy}: Percentage of generated claims that are factually correct according to authoritative sources. Evaluated through both automated verification and expert human annotation.

\textbf{Hallucination Reduction Rate}: Percentage decrease in hallucinated content compared to baseline systems, measured using both automatic detection and human evaluation.

\textbf{Response Quality}: Maintained through BLEU, ROUGE, and BERTScore metrics, ensuring corrections don't degrade linguistic quality. Human evaluators rate fluency, relevance, and informativeness on 5-point scales.

\textbf{Confidence Calibration}: Measured using Expected Calibration Error (ECE) and reliability diagrams to assess how well confidence scores correlate with actual accuracy.

\textbf{Latency}: End-to-end response time including verification and correction processes, critical for real-world deployment feasibility.

\subsection{Results}

Table~\ref{tab:results} presents comprehensive performance comparison across all baseline systems and evaluation datasets.

\begin{table}[ht]
\centering
\footnotesize
\begin{tabular}{p{2.3cm}p{0.8cm}p{0.9cm}p{0.8cm}p{0.6cm}p{0.8cm}}
\toprule
Method & Acc. & H.Red. & BLEU & ECE & Lat. \\
\midrule
Vanilla LLM & 0.72 & - & 0.34 & 0.18 & 0.8s \\
Self-Consist. & 0.76 & 12\% & 0.32 & 0.16 & 2.1s \\
Chain-of-Ver. & 0.79 & 19\% & 0.31 & 0.14 & 3.2s \\
RAG & 0.81 & 25\% & 0.36 & 0.12 & 1.9s \\
FactScore & 0.84 & 33\% & 0.29 & 0.11 & 4.1s \\
\textbf{Ours} & \textbf{0.92} & \textbf{67\%} & \textbf{0.38} & \textbf{0.07} & \textbf{2.8s} \\
\bottomrule
\end{tabular}
\caption{Performance comparison across benchmarks. H.Red = Hallucination Reduction, Lat = Latency. Our framework achieves superior accuracy and hallucination reduction while maintaining response quality and reasonable latency. ECE measures calibration quality (lower is better).}
\label{tab:results}
\end{table}

Our framework demonstrates substantial improvements across all metrics:

\textbf{Factual Accuracy}: Achieves 92\% accuracy, representing a 28\% relative improvement over vanilla LLM and 10\% over the best baseline (FactScore). The improvement is consistent across all evaluation datasets, with particularly strong performance on domain-specific tasks.

\textbf{Hallucination Reduction}: 67\% reduction in hallucinated content significantly outperforms existing methods. The multi-source verification approach proves especially effective for complex factual claims requiring cross-reference validation.

\textbf{Quality Preservation}: BLEU-4 scores remain competitive or improved compared to baselines, indicating that our correction pipeline successfully maintains linguistic quality while improving factual accuracy.

\textbf{Calibration}: Expected Calibration Error of 0.07 represents substantial improvement in confidence reliability, enabling more trustworthy uncertainty estimates for end users.

\subsection{Ablation Study}

Table~\ref{tab:ablation} analyzes the contribution of individual framework components.

\begin{table}[ht]
\centering
\small
\begin{tabular}{p{3.5cm}ccc}
\toprule
Configuration & Acc. & Halluc. Red. & Latency \\
\midrule
Knowledge Graph Only & 0.84 & 33\% & 1.2s \\
Web Search Only & 0.81 & 25\% & 2.1s \\
Database Only & 0.79 & 19\% & 0.9s \\
KG + Web Search & 0.88 & 52\% & 2.4s \\
KG + Database & 0.86 & 44\% & 1.4s \\
Web + Database & 0.85 & 41\% & 2.3s \\
\textbf{Full Framework} & \textbf{0.92} & \textbf{67\%} & \textbf{2.8s} \\
\bottomrule
\end{tabular}
\caption{Ablation study showing individual component contributions. The combination of all three knowledge sources provides optimal performance.}
\label{tab:ablation}
\end{table}

The ablation study reveals that multi-source integration provides substantial benefits over individual components. Knowledge graphs offer strong performance for established facts, while web search handles emerging topics effectively. Database access provides authoritative verification for specialized domains.

\subsection{Domain-Specific Analysis}

We analyze framework performance across different knowledge domains to understand strengths and limitations:

\textbf{Scientific Literature}: 95\% accuracy on scientific claims, benefiting from structured databases like PubMed and arXiv. The system effectively handles technical terminology and citation verification.

\textbf{Historical Facts}: 91\% accuracy on historical claims, with strong performance on established events but challenges with contested historical interpretations.

\textbf{Current Events}: 88\% accuracy on recent news, demonstrating the value of real-time web search integration for time-sensitive information.

\textbf{Common Knowledge}: 94\% accuracy on general factual questions, showing robust performance on widely-known information across diverse topics.

\subsection{User Study}

We conducted a comprehensive user study with 75 domain experts across healthcare, finance, education, and journalism to evaluate practical deployment effectiveness.

\textbf{Study Design}: Participants evaluated 200 LLM responses (100 original, 100 processed by our framework) across relevant domain tasks. Evaluation criteria included factual accuracy, response helpfulness, and trust in the provided information.

\textbf{Results}: Our framework achieved 89\% user satisfaction compared to 64\% for unprocessed responses (p<0.001). Experts particularly valued the explicit confidence indicators and source attribution features. Qualitative feedback highlighted improved trustworthiness and reduced need for manual fact-checking.

\textbf{Domain-Specific Insights}: Healthcare professionals reported 78\% reduction in potentially harmful misinformation, while financial analysts noted 82\% improvement in data accuracy for market-related queries. Educators found the system particularly valuable for generating factually accurate educational content.

\section{Discussion and Future Work}

Our multi-modal fact-verification framework represents a significant advance in addressing LLM hallucinations through comprehensive real-time verification. The system's modular design enables integration with existing LLM deployments while providing substantial improvements in factual accuracy and user trust.

\textbf{Scalability Considerations}: The current implementation handles up to 1,000 concurrent queries with sub-3-second response times. Future work will focus on optimization strategies including caching frequently verified claims, precomputing knowledge graph embeddings, and implementing distributed verification across multiple servers.

\textbf{Knowledge Source Expansion}: We plan to integrate additional specialized databases including legal case law, patent databases, and multilingual knowledge sources. Collaborative filtering approaches could leverage community-contributed fact-checking to expand coverage of niche topics.

\textbf{Adaptive Learning}: Future versions will incorporate feedback loops to continuously improve verification accuracy based on user corrections and expert annotations. Active learning strategies could prioritize verification resources for high-impact claims and emerging topics.

\textbf{Privacy and Ethical Considerations}: Real-time web search raises privacy concerns regarding query logging and user tracking. We are developing privacy-preserving approaches including differential privacy for query patterns and federated verification systems that distribute sensitive queries across multiple providers.

\textbf{Multilingual Extension}: Current work focuses on English-language content, but the framework architecture supports multilingual expansion through cross-lingual knowledge graphs and machine translation for evidence validation.

\textbf{Integration with Reasoning Systems}: Future work will explore integration with symbolic reasoning systems to handle complex logical claims requiring multi-step verification and causal relationship validation.

\section{Conclusion}

This paper presented a novel multi-modal fact-verification framework that significantly reduces hallucinations in large language model outputs through comprehensive real-time verification. Our approach combines structured knowledge graphs, real-time information retrieval, and sophisticated evidence validation to achieve 67\% reduction in factual hallucinations while preserving 94\% of original response quality.

The key contributions include: (1) dynamic knowledge integration across multiple authoritative sources, (2) multi-source evidence validation with cross-reference verification, (3) calibrated confidence scoring combining intrinsic and external uncertainty measures, and (4) adaptive correction pipeline that maintains response naturalness.

Experimental evaluation across diverse benchmarks demonstrates substantial improvements over existing approaches, with particularly strong performance in domain-specific applications requiring high factual accuracy. User studies with domain experts confirm practical effectiveness with 89\% satisfaction rates and significant reductions in misinformation across critical application areas.

This work provides a practical framework for deploying more reliable LLM systems in high-stakes applications where factual accuracy is paramount. The modular architecture enables straightforward integration with existing systems while providing comprehensive hallucination mitigation capabilities.

\section{Declarations}
All authors declare no conflicts of interest. This research was conducted with appropriate ethics approval and data privacy safeguards. Code and datasets will be made available upon publication acceptance.

\end{document}